\documentclass[11pt,a4paper]{article}
\usepackage[hyperref]{naaclhlt2018}
\usepackage{times}
\usepackage{latexsym}
\usepackage{graphicx}
\usepackage{enumitem}
\usepackage{flexisym}
\usepackage{booktabs}       
\usepackage{textcomp}
\usepackage{float}
\usepackage{multirow}
\usepackage{url}
\usepackage{amsfonts}
\newcommand{\bq}{\mathbf{q}}
\newcommand{\br}{\mathbf{r}}

\aclfinalcopy 

\title{Knowledge Base Relation Detection via Multi-View Matching}
\author{
  Yang Yu$^\dagger$ \\
  \texttt{yu@us.ibm.com} \\
  \And
  Kazi Saidul Hasan$^{\dagger}$\\
  \texttt{kshasan@us.ibm.com} \\
  \AND
  Mo Yu$^\ddagger$ \quad\quad\quad\quad\quad Wei Zhang$^\ddagger$\quad\quad\quad\quad\quad Zhiguo Wang$^\S$\\
  \texttt{yum@us.ibm.com}\quad  \texttt{zhangwei@us.ibm.com}\quad \texttt{zhigwang@us.ibm.com}\\
  \\
  $^\dagger$IBM Watson, USA\\
  $^\ddagger$AI Foundations, IBM Research, USA\\
  $^\S$IBM Research, USA
}

\date{}

\begin{document}
\maketitle
\begin{abstract}
Relation detection is a core component for Knowledge Base Question Answering (KBQA). In this paper, we propose a knowledge base (KB) relation detection model based on multi-view matching, which utilizes useful information extracted from questions and KB.%
The matching inside each view is through multiple perspectives to compare two input texts thoroughly. All these components are trained in an end-to-end neural network model. Experiments on SimpleQuestions and WebQSP yield state-of-the-art results on relation detection.
\end{abstract}

\section{Introduction}

Knowledge Base Question Answering (KBQA) systems query a knowledge base (KB) (e.g., Freebase, DBpedia) to answer questions \cite{berant2013semantic,yao2014freebase,bordes2015large,bast2015more,yih2015semantic,xu2016enhancing}.
To transform a natural language question to a KB query, a KBQA system needs to perform at least two sub-tasks: (1) detect KB entities appearing in a question and (2) detect a KB relation associated with a question. 
A KBQA system usually includes separate components to accomplish these sub-tasks.
This paper focuses on the second sub-task, frequently referred to as \textbf{relation detection}, to identify which KB relation(s) are expressed by a given question.
As discussed in Yu et al.~\shortcite{KBQA-yum2017}, relation detection remains a bottleneck of KBQA systems owing to its inherent difficulty.

KB relation detection is more challenging compared to general relation detection \cite{zhou_exploring_2005,rink-harabagiu:2010:SemEval,sun_semi-supervised_2011,nguyen_employing_2014,gormley-yu-dredze:2015:EMNLP,nguyen2015combining} in the field of IE for two key reasons: (1) the number of relations to predict is usually large ($>$1k) and (2) relations in a test set may not be seen during training.
Previous work mainly focused on the large relation vocabulary problem and zero-shot relation learning. For example, Dai et al.~\shortcite{dai-li-xu:2016:P16-1} use pre-trained relation embeddings from TransE \cite{bordes2013translating} to initialize the relation representations. Yin et al.~\shortcite{yin2016simple} and Liang et al.~\cite{liang2016neural} factorize relation names into word sequences motivated by the fact that KB relation names usually comprise meaningful word sequences.
Yih et al.~\shortcite{yih2015semantic} and Golub et al.~\shortcite{golub2016character} represent questions at character level. 
Yu et al.~\shortcite{KBQA-yum2017} propose to use both granularities of relation names and words in a hierarchical encoding and matching framework.

In this paper, we propose to improve KB relation detection by exploiting multiple views i.e., by leveraging more information from KB to obtain better question-relation matching. Besides frequently used relation names, we propose to make use of entity type(s) a relation can logically have as objects (i.e., object in a KB triple $<$$subject, predicate, object$$>$). For instance, for a given question ``\emph{What country is located in the Balkan Peninsula?}'', the correct relation is $contains$ and the object type for this relation is $location$. We hypothesize that, in addition to relation names, it may also be useful to match this question against the object entity type (i.e., $location$) since the question has the word ``located'', indicating that the answer to this question is a location.      

Our contributions are two-fold. (1) We formulate relation detection as a multi-view matching task, where multiple views of information from both question and relation are extracted. We use an attention-based model to compare question and relation from multiple perspectives in each view. (2) We exploit object entity types, automatically extracted from KB, in our multi-view matching model. These two contributions help us achieve state-of-the-art KB relation detection accuracies on both WebQSP and SimpleQuestions datasets.

\section{Related Work}
\label{sec:relatedwork}

\paragraph{Relation Extraction}
Relation extraction (RE) was researched originally as an sub-field of information extraction. The major research methods in the traditional RE has the knowledge of a (small) pre-defined relation set, then given a text sequence and two target entities, the goal of these methods is to choose a relation or none which means if this relation or no relation holds between the two target entities. Thus from another perspective, RE methods are usually described as a \textbf{classification task}.
Most of these RE methods need a step to manually pick large amount of features\cite{sun_semi-supervised_2011,zhou_exploring_2005,rink-harabagiu:2010:SemEval}.
Due to recent machine learning and especially deep learning advances, many recent proposed RE approaches begin to explore the benefits of deep learning instead of using hand-crafted features.
The main benefits ranging from pre-trained word embeddings \cite{gormley-yu-dredze:2015:EMNLP,nguyen_employing_2014} to deep neural networks like convolutional neural networks (CNN) and long-short term memories (LSTMs) \cite{vu-EtAl:2016:N16-1,zeng-EtAl:2014:Coling,santos2015classifying} and attention models \cite{zhou-EtAl:2016:P16-2,wang-EtAl:2016:P16-12} which is shown to be key for a lot of other NLP tasks, such as machine translation, named entity recognition, reading comprehension, etc.

One strong assumption mentioned above in the most RE methods is that a fixed (i.e., closed) set of relation types is given as an prior knowledge, thus no zero-shot learning capability (i.e. detecting new relations that did not occur during training) is required. Another commonality among these RE methods is that the relation set is usually not large. Here are some examples. The widely used ACE2005 has 11/32 coarse/fine-grained relations; SemEval2010 Task8 has 19 relations; TAC-KBP2015 has 74 relations although it considers open-domain Wikipedia relations. Compared to that, KBQA usually has thousands of relations. Thus most RE approaches may not work well by directly being adapted to large number of relations or unseen relations.
The relation embeddings in a low-rank tensor method were used \cite{yu-EtAl:2016:N16-12}. However it is still using supervised way to train their relation embeddings and relation set used in the experiments is still not large.

\paragraph{Relation Detection in KBQA Systems}
Similar to how RE methods evolved over time, relation detection methods for KBQA were also originally based on many hand-crafted features \cite{bast2015more,yao2014information}. Later researchers started to explore the benefits of some simple deep neural networks \cite{dai-li-xu:2016:P16-1,yih2015semantic,xu2016enhancing} and some advances onces including attention models \cite{golub2016character,yin2016simple}.

In order to work well for open-domain question answering, many of the relation detection research for KBQA are designed to support large relation set and even open relation sets, for example ParaLex \cite{fader2013paraphrase}) and SimpleQuestions which are datasets need the capacity to support large relation sets and unseen relations becomes more necessary. While some KBQA data does not take such abilities into consideration because of the unnatural distribution of testing data: most of the gold test relations can be observed during training. For example WebQuestions, such property makes it less a open-domain task, thus the problem of supporting full relation vocabulary and zero-shot learning becomes less serious. Thus some prior work on this task adopted the close domain assumption like in the general RE research. 

To support open QA, there are two main solutions for relation detection: (1) use pre-trained relation embeddings (e.g. from TransE \cite{bordes2013translating}), like \cite{dai-li-xu:2016:P16-1}; (2) factorize the relation names to sequences and formulate relation detection as a \textbf{sequence matching and ranking} task. Such factorization works because that the relation names usually comprise meaningful word sequences, especially for the OpenIE patterns such as in ParaLex.
For example, relations are split into word sequences for single-relation detection \cite{yin2016simple}. Also good performance was achieved on WebQSP with word-level relation representation in an end-to-end neural programmer model \cite{liang2016neural} .
Character tri-grams was used as inputs on both question and relation sides \cite{yih2015semantic}. \newcite{golub2016character} proposed a generative framework for single-relation KBQA which predicts relation with a character-level sequence-to-sequence model.

Another significant difference between relation detection in KBQA and general RE is that general RE research works on the condition that the two argument entities are both available. Thus it usually can learn from features \cite{gormley-yu-dredze:2015:EMNLP,nguyen_employing_2014} or attention mechanisms \cite{wang-EtAl:2016:P16-12} based on the entity information (e.g. entity types or entity embeddings).
In contrast relation detection for KBQA mostly does not have this information: (1) one question usually contains single argument (the topic entity) and (2) one KB entity could have multiple types (type vocabulary size larger than 1,500). This makes KB entity typing itself a difficult problem so no previous used entity information in the relation detection model. Such entity information has been used in some KBQA systems as features for the final answer re-rankers.

\section{Problem Overview}

\paragraph{Problem Definition}
Formally, for an input question $\bq$, the task is to identify the correct relation $\br^{(gold)}$ from a set of candidate relations $\mathcal{R}=\{\br\}$. The problem thus becomes learning a scoring function $s(\br|\bq)$ for optimizing some ranking loss.

Both questions and relations have different views of input features. Each view can be written as a sequence of tokens (regular words or relation names). Therefore, for a view $i$ of relation $\br$, we have $\br^{(i)}=\{r^{(i)}_1,\cdots,r^{(i)}_{M_i}\}$, where $M_i$ is the length of relation $\br$'s word sequence for view $i$.
The same definition holds for the question side. Finally, we have the multi-view inputs for both a question $\bq=\{\bq^{(1)},\cdots,\bq^{(N_q)}\}$ and a relation $\br=\{\br^{(1)},\cdots,\br^{(N_r)}\}$, where $N_q$ and $N_r$ denote the number of views for $\bq$ and $\br$, respectively. Note that $N_q$ and $N_r$ may not be equal.

\paragraph{Views for KB Relation Detection}
For an input question, we generate views from relation names and their corresponding tail entity types and use three pairs of inputs in the model (see Figure~\ref{fig:arch}).

\begin{enumerate}[labelindent=0pt,labelwidth=0.75em,leftmargin=!]
\item \emph{$<$entity name, entity mention$>$} pair captures entity information from question and KB.
\vspace{-2mm}
\item \emph{$<$relation name, abstracted question$>$} pair captures the interaction between an input question and a candidate relation. Following previous work \cite{yin2016simple,KBQA-yum2017}, we replace the entity mention in a question by a special token (``Balkan Peninsula'' is replaced by $<$e$>$ in Figure~\ref{fig:arch}) to become an abstracted question, so that the model could focus better on matching a candidate relation name to the entity's context in a question.
\vspace{-2mm}
\item \emph{$<$relation tail entity types, abstracted question$>$} pair helps determine how well relation tail types match with a question. Section~\ref{section:relation_tail_types} describes how we extract and use tail entity types.
\end{enumerate}

For an input question, the first pair of inputs remains the same for all candidate relations to help the model differentiate between the candidates. So this pair does not need to be thoroughly compared via multi-perspective matching as all the other pairs of inputs.

For inputs to the 2nd and 3rd view, we generate two matching feature vectors, one for each of the directions of matching (i.e., for a pair $<$$a,b$$>$, the directions are $a$$\rightarrow$$b$ and $a$$\leftarrow$$b$). Finally, the model combines these two pairs of interaction information to have a high-level joint view. The joint view helps us detect the most promising relation given how the question matches with the candidate relation names and the corresponding tail entity types. We present more details in Section~\ref{Section:mvm} and present the experimental results with different combinations of views in Section~\ref{sec:exp}.  


\section{Relation Tail Entity Types Extraction}
\label{section:relation_tail_types}

In this work, we propose to make use of entity type(s) a relation can logically have as tails (i.e., object in a KB triple $<$$subject, predicate, object$$>$). More often than not, KB relations can only have tail entities of specific types. For instance, for our example question ``\emph{What country is located in the Balkan Peninsula?}'', the corresponding relation in Freebase is $contains$ and the tail entity (i.e., the answer to the question) can only be of type $location$. This and other relations such as $adjoin\_s$, $street\_address$, $nearby\_airports$, $people\_born\_here$ can only have locations as tail entities, however the relations do not explicitly contain word(s) indicating the type of entities expected as answers. Motivated by this, we hypothesize that exploiting tail entity type information may improve relation detection performance. For our example, the learner may exploit the tail entity type (i.e., $location$) to learn that the relations are somewhat similar as they all share the same tail entity type and learn more generic representations for relations that have locations as tail types. Yin et al.~\cite{Type-aware-KBQA} also exploit tail entity types as they predict answer entity type as an intermediate step before predicting an answer. In contrast, we describe next how we heuristically generate a short list of relevant tail entity types for each unique KB relation. 

A tail entity in an instance of a relation may be associated with multiple types. Given the triple $<$$The\_Audacity\_of\_Hope$, $author$, $Barack\_Obama$$>$, $Barack\_Obama$ has types ranging from as generic as $person$ to more specific ones such as $writer$, $politician$, and $us\_president$. Therefore, given the relation $author$, it is crucial to prune the unrelated entity types ($politician$, $us\_president$) and retain the relevant ones ($person$, $writer$). To achieve this, we first obtain at most 500 instances
\footnote{{We empirically found that 500 instances were sufficient for our entity type extraction experiment.}} for each unique relation from Freebase. Next, we query for the types for each of the tail entities obtained in the first step.\footnote{{In Freebase, the relation $type.object.type$ lists the types for an entity.}} Finally, we retain only the types that at least 95\% of the tail entities have. A default special token is used if we can not find any tail entity type for a relation in this approach. Once the tail types are obtained for a particular relation, we form one string by concatenating the words in each of the tail types and use the string as tail entity type string in the model described in Section~\ref{Section:mvm}. 

\section{Model Architecture}
\label{Section:mvm}

\begin{figure*}[h]
  \centering
\includegraphics[scale=0.3]{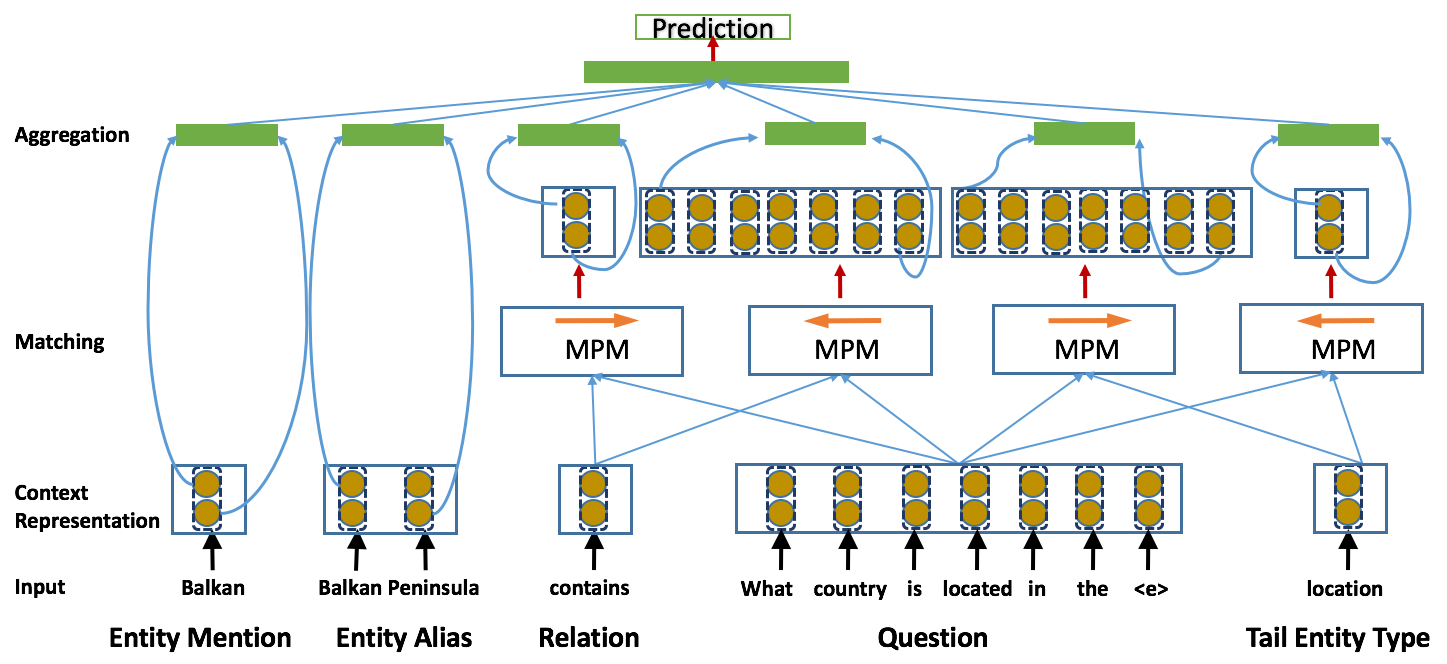}
  \caption{Multiple view matching for detecting relation in ``What country is located in the Balkan Peninsula?''.}
\label{fig:arch}
\end{figure*}

Figure \ref{fig:arch} illustrates the architecture of our model. Apart from the entity alias and entity span pair (henceforth referred to as \emph{entity pair}), each pair of inputs is matched from multiple perspectives, and then the matching representations of all pairs and the representations of entity pair are aggregated for final prediction. Next, we describe the three main components: inputs, context representation module, matching module and aggregation module.

\paragraph{Input Module}
The inputs to all views in the model are word sequences, and our model encodes each sequence in two steps. First, the model constructs a $d$-dimensional vector for each word with two components: a word and a character-based embedding. A word embedding is a fixed, pre-trained vector (e.g., GloVe \cite{pennington2014glove}, word2vec \cite{mikolov2013distributed}). A character-based embedding is calculated by feeding each character (also represented by a vector) within a word into a LSTM.

\paragraph{Context Representation Module}
The model leverages the same BiLSTM to encode all views of inputs. Then, the output contextual vectors of each BiLSTM are used in the matching modules. The contextual vectors for a question are fed into multiple matching modules to match with relation and tail types.

The purpose of this module is to incorporate contextual information into the representation of each time step of each input sequence. We utilize a bi-directional LSTM (BiLSTM) to encode contextual embeddings for each time-step of input sequence and get hidden state for each word position. We can have separated parameters for question and passage encoders but a single shared encoder for both works better in the experiments.

As RNN input, a word is represented by a row vector $x \in \mathbb{R}^n$. $x$ can be the concatenation of word embedding and word features, though we do not use any additional word features. The word vector for the $t$-th word is $x_t$. A word sequence is processed using an RNN encoder with long-short term memory (LSTM) \cite{LSTM}, which was proved to be effective in many types of NLP tasks, including machine reading comprehension and neural machine translation tasks \cite{bahdanau2014neural,kadlec2016text,dhingra2016gated}. For each position $t$, LSTM computes $h_t$ with input $x_t$ and previous state $h_{t-1}$, as:
\begin{eqnarray}
  i_{t} &=&\sigma(W_{i}x_{t}+U_{i}h_{t-1}) \\
  f_{t} &=&\sigma(W_{f}x_{t}+U_{f}h_{t-1}) \\
  o_{t} &=&\sigma(W_{o}x_{t}+U_{o} h_{t-1}) \\
  \tilde{c}_{t}&=&tanh(W_{c}x_{t}+U_{c}h_{t-1}) \\
  c_{t} &=&f_{t}\cdot \tilde{c}_{t-1}+ i_{t} \cdot \tilde{c}_{t} \\
  h_{t} &=&o_{t}\cdot tanh(c_{t})
\end{eqnarray}
where $h_t$, $i_t$, $f_t$, $o_t$ and $c_t \in \mathbb{R}^d$ are d-dimensional hidden state, input gate, forget gate, output gate and memory cell, respectively; $W_{\{i,f,o,c\}}$, $W \in \mathbb{R}^{n\times d}$ and $U_{\{i,f,o,c\}}$, $U \in \mathbb{R}^{d\times d}$ are the parameters of the LSTM; $\sigma$ is the sigmoid function, and $\odot$ denotes element-wise production. For a word at $t$, we use the hidden state $\overrightarrow{h}_t$ from the forward RNN as a representation of the preceding context, and the $\overleftarrow{h}_t$ from a backward RNN that encodes text reversely, to incorporate the context after $t$. Next, $h_t=[\overrightarrow{h_t};\overleftarrow{h_t}]$, the bi-directional contextual encoding of $x_t$, is formed. $[\cdot ;\cdot]$ is the concatenation operator. To distinguish hidden states from different sources, we denote the $h_j$ of $j$-th word in $P$ and the $h_k$ of $k$-th word in $Q$ as $h^p_{j}$ and $h^q_{k}$ respectively.
\begin{eqnarray}
\overrightarrow{h_i} = \overrightarrow{LSTM}(\overrightarrow{h}_{i-1}, p_i) \quad\quad i = 1,...,M \\
\overleftarrow{h_i} = \overleftarrow{LSTM}(\overleftarrow{h}_{i+1}, p_i) \quad\quad i = M,...,1
\end{eqnarray}
\paragraph{Matching Module}
The core task of relation detection is to calculate information interaction between relations and the given question.
In this work, we design the matching module with attention models to match each view of a relation with a given question. The reason attention could be important here is that different views of relations usually correspond to different parts of questions. For example in Figure \ref{fig:arch}, the question words are usually more likely to indicate the relation types. 

We modify the bilateral multiple perspective matching (BiMPM) model~\cite{MPM_ijcai2017}, which performs comparably with state-of-the-art systems for several text matching tasks. We hypothesize that BiMPM could also be effective for relation detection since a unique view of a question may be required to match with either a relation or a tail entity type, and the matching method should match the relation with the question in multiple granularities and multiple perspectives. 


In Figure~\ref{fig:arch}, each box at the ``Matching'' layer is a single directional multi-perspective matching (MPM) module, therefore two such boxes together form a BiMPM module. \textbf{We have modules on all views share the same parameters in the experiments.}. Each MPM module takes two sequences, an anchor and a target, as inputs, and matches each contextual vector of the anchor with all the contextual vectors of the target. The arrows inside the MPM boxes in Figure~\ref{fig:arch} denote the direction of matching i.e., anchor $\rightarrow$ target. To form a BiMPM, for instance, a question and a relation are considered anchor and target, respectively, and vice versa. During matching, a matching vector is calculated for each contextual vector of the anchor by composing all the contextual vectors of the target. Then, the model calculates similarities between the anchor contextual vector and the matching vector from multiple perspectives using the multi-perspective cosine similarity function.

The multiple-perspective cosine matching function $f_m$ to compare two vectors is 
\begin{equation}
\label{eq_m}
m=f_m(v_1,v_2; W)
\end{equation}
In the equation \ref{eq_m}, $v_1$ and $v_2$ are two d-dimensional vectors, $W\in R^{l\times d}$ is a trainable parameter with the shape $l\times d$, $l$ is the number of perspectives, and the returned value $m$ is a l-dimensional vector. Each element $m_k\in m$ is a matching value from the k-th perspective, and it is calculated by the cosine similarity between two weighted vectors 
\begin{equation}
m_k=cosine(W_k\odot v_1, W_k\odot v_2)
\end{equation}
where $\odot$ is the element-wise multiplication, and $W_k$ is the k-th row of $W$, which controls the k-th perspective and assigns different weights to different dimensions of the d-dimensional space.

The MPM module uses four matching strategies in this regard. 

(1) \textbf{Full-Matching:} Each contextual vector of an anchor is compared with the last contextual vector of a target, which represents the entire target sequence.
\begin{eqnarray}
\overrightarrow{m}_i^{full} = f_m(\overrightarrow{h}_i^{anchor},\overrightarrow{h}_N^{target};W^1) \nonumber \\
\overleftarrow{m}_i^{full} = f_m(\overleftarrow{h}_i^{anchor},\overleftarrow{h}_1^{target};W^2) \nonumber
\end{eqnarray}

(2) \textbf{Max-Pooling-Matching:} Each contextual vector of an anchor is compared with every contextual vector of the target with the multi-perspective cosine similarity function, and only the maximum value of each dimension is retained.
\begin{eqnarray}
\overrightarrow{m}_i^{max} = \max\limits_{j\in(1...N)}f_m(\overrightarrow{h}_i^{anchor},\overrightarrow{h}_j^{target};W^3)  \nonumber\\
\overrightarrow{m}_i^{max} = \max\limits_{j\in(1...N)}f_m(\overleftarrow{h}_i^{anchor},\overleftarrow{h}_j^{target};W^4) \nonumber
\end{eqnarray}
where $\max\limits_{j\in(1...N)}$ is element-wise maximum.

(3) \textbf{Attentive-Matching:} First, the cosine similarities between all pairs of contextual vectors in the two sequences are calculated. Then the matching vector is calculated by taking the weighted sum of all contextual vectors of the target, where the weights are the cosine similarities computed above.
\begin{eqnarray}
\overrightarrow{\alpha}_{i,j} = cosine(\overrightarrow{h}_i^{anchor},\overrightarrow{h}_j^{target}) \quad j=1,...,N \nonumber \\
\overleftarrow{\alpha}_{i,j} = cosine(\overleftarrow{h}_i^{anchor},\overleftarrow{h}_j^{target}) \quad j=1,...,N \nonumber
\end{eqnarray}

\begin{eqnarray}
\overrightarrow{h}_i^{mean}=\dfrac{\sum^N_{j=1}{\overrightarrow{\alpha}_{i,j} \cdot \overrightarrow{h}_j^{target}}}{\sum^N_{j=1}{\overrightarrow{\alpha}_{i,j}}}  \nonumber\\
\overleftarrow{h}_i^{mean}=\dfrac{\sum^N_{j=1}{\overleftarrow{\alpha}_{i,j} \cdot \overleftarrow{h}_j^{target}}}{\sum^N_{j=1}{\overleftarrow{\alpha}_{i,j}}} \nonumber
\end{eqnarray}
\begin{eqnarray}
\overrightarrow{m}_i^{att} = f_m(\overrightarrow{h}_i^{anchor},\overrightarrow{h}_i^{mean};W^5) \nonumber \\
\overleftarrow{m}_i^{att} = f_m(\overleftarrow{h}_i^{anchor},\overleftarrow{h}_i^{mean};W^6) \nonumber
\end{eqnarray}
(4) \textbf{Max-Attentive-Matching:} This strategy is similar to Attentive-Matching except that, instead of taking the weighted sum of all the contextual vectors as the matching vector, it picks the contextual vector with the maximum cosine similarity from the target.

\paragraph{Aggregation Module}
The first step in this module is to apply another BiLSTM on the two sequences of matching vectors individually. Then, we construct a fixed-length matching vector by concatenating vectors from the last time-step of the BiLSTM models. This is the representation of the overall matching for one view.

For combining the matching results from different views of input pairs and entity pair, we have the aggregation layer at the end, which takes the matching representations or scores from different views and extracted feature representation for entity pair, then constructs a feature vector for relation prediction. In this work, we simply use the concatenation of different matching representations generated from all the views by the matching modules. The combined representation of all multiple views are transformed into a final prediction through a multiple perception layer.


\section{Experiments}
\label{sec:exp}

\begin{table*}[t]
  \centering
  \begin{tabular}{ccccc}
    \toprule
Row&Model&Char&WQ&SQ\\
\hline
1 & BiCNN \cite{yih2015semantic} & Y & 77.74 &90.0 \\
2 & AMPCNN \cite{yin2016simple} & N/A & - &91.3\\
3 & Hier-Res-BiLSTM \cite{KBQA-yum2017} & N/A & 82.53 &93.3\\
\hline
4 & (Q\textprime, Relation)&Y&75.26&93.13\\
5 & (Q\textprime, Relation)&N&75.63&93.25\\
6 & (Q\textprime, Relation+Type)&Y&76.41&93.29\\
7 & (Q\textprime, Relation+Type)&N&75.95&93.43\\
8 & (Q, Relation)(Q, Type)&Y&83.71&93.13\\
9 & (Q\textprime, Relation)(Q\textprime, Type)&Y&84.74&93.38\\
10 & (Q\textprime, Relation)(Q\textprime, Type)&N&84.86&93.52\\
11 & (Entity Pair)(Q\textprime, Relation)(Q\textprime, Type)&Y&\textbf{85.95}&93.69\\
12 & (Entity Pair)(Q\textprime, Relation)(Q\textprime, Type)&N&{85.41}&\textbf{93.75}\\
\bottomrule
  \end{tabular}
  \caption{Relation detection accuracies for WQ and SQ. The second column lists the pairs of inputs (enclosed in parentheses) matched in our model. Q and Q\textprime denote original and abstracted (i.e., entity mention replaced) question text, respectively. ``Relation'' and  ``Type'' denote candidate relation and tail entity types text, respectively. ``Relation+Type'' denotes a single input, where relation and  tail entity types are concatenated by a special symbol. ``Entity Pair'' refers to entity alias and entity span pair. ``Char'' column shows if character embeddings are used besides word embeddings.}
  \label{tbl:main_results}
\end{table*}

\paragraph{Datasets} We use two standard datasets - SimpleQuestions (SQ) \cite{bordes2015large} and WebQSP (WQ) \cite{yih-EtAl:2016:P16-2}. Each question in these datasets is labeled with head entity and relation information. SQ has only single-relation questions i.e., there is one $<$$head$, $relation$, $tail$$>$ triple per question. In contrast, WQ has both single and multiple-relation questions. For a multiple-relation question, there are multiple relations on the path connecting a head to a tail entity. We adopt the same approach as Yu et al.~\shortcite{KBQA-yum2017} to create positive and negative instances.


\textbf{SimpleQuestions (SQ): } The dataset has only single-relation questions i.e., a head entity and a tail entity is connected by one relation. To compare with previous work \cite{bordes2015large}, we use a subset of Freebase with 2M entities (FB2M). We use the same training, validation, and test sets used in \cite{bordes2015large}.

It is a single-relation KBQA task. The KB we use consists of a Freebase subset with 2M entities (FB2M) \cite{bordes2015large}, in order to compare with previous research. \cite{yin2016simple} also evaluated their relation extractor on this data set and released their proposed question-relation pairs, so we run our relation detection model on their data set. The training set has 571k instances and each question on average has about 8 candidate relations.

\textbf{WebQSP (WQ): } Unlike SimpleQuestions, WebQSP has both single and multi-relation questions. In case of a multi-relation, a head and a tail entity are connected via one or more contextual vectorTs (Compound Value Type) and there are multiple relations on the path connecting the head entity, the contextual vectorT(s), and the tail entity. We use the entire Freebase for our experiments on this dataset. The train and test sets are the same as the ones used in \cite{yih-EtAl:2016:P16-2}. We adopt the same approach as Yu et al.~\cite{KBQA-yum2017} to create positive and negative instances.

A multi-relation KBQA task. We use the entire Freebase KB for evaluation purposes. Following \cite{KBQA-yum2017}, we evaluate the relation detection models through a new relation detection task from the WebQSP data set.
The training set has 215k instances and each question on average has about 71 candidate relations.

\paragraph{Experimental Setup} We used development sets to pick the following hyper-parameter values: (1) the size of hidden states for LSTMs (300); (2) learning rate (0.0001); and (3) the number of training epochs (30). 
All word vectors are initialized with 300-$d$ GloVe embeddings~\cite{pennington2014glove}. During testing, we predict the candidate relation with the highest confidence score.


\paragraph{Results and Analysis}
Table~\ref{tbl:main_results} shows that our model yields state-of-the-art relation detection scores for both WQ (Row~11) and SQ (Row~12) by beating the previous best system \cite{KBQA-yum2017} by 3.42 and 0.45 points, respectively. 

Rows~8-10 show that using relation and tail type as two separate inputs consistently outperforms the setting, where they are provided as a single input (Rows~6-7). Rows~8-9 also show that replacing entity mentions in question texts helps our model to focus more on the contextual parts of questions.

We found that using character embeddings on top of word embeddings does not have any significant impact. We hypothesize that this is due to the small number of KB relations and tail types. Although there are several thousands of these in Freebase, they are still much smaller in number compared to a vocabulary obtained from a large text corpus. Owing to this, there is little scope for character embeddings to capture prefix, suffix, or stem patterns that can otherwise be observed more frequently in a large corpus. 

As the scores indicate, WQ is more difficult than SQ and several reasons may contribute to this trend. First, owing to multi-relations, the average number of candidate relations per question is more in WQ. Second, WQ has more questions that are close to real world questions asked by humans. In contrast, the questions in SQ are synthetic in nature as they are composed by looking at the true answer in KB. Third, WQ needs more complex reasoning on KB, as the path from head entity to answer often consists of multiple hops. As a result, scores for SQ are in the 90s whereas there is still room for improvement for WQ. 

Last two rows show that our proposed model achieves the best performance on both WQ and SQ. While replacing entity mentions yields improvement, the model cannot use entity information in this process. However, our results confirmed that extracting features from entity pair inputs separately for final prediction was useful.

As the multi-perspective matching using question with entity mention replaced, it helps the model to focus on what information needed. However it also reduces information about entity, so we think extracting features from entity pair for final prediction should help and the results confirmed it.

From the table \ref{tbl:main_results}, we see that adding relation tails entity types always helps. If using it more appropriately by matching in different view with question, it can helps significantly. Comparing row 8 and 9, using question text with entity mention being replaced helps model to focus on important information from question than using original question text.

From the table \ref{tbl:main_results}, we can also see that adding character embedding on top of word embedding is not that helpful. We think the main reason is the vocabulary is limited in KBQA scenario than normal texts. The word variation is less, thus character embedding could not capture a lot prefix/postfix/stemming that could helps in other NLP tasks.


\section{Conclusion}
Relation detection, a crucial step in KBQA, is significantly different from general relation extraction. To accomplish this task, we propose a novel KB relation detection model that performs bilateral multiple perspective matching between multiple views of question and KB relation. Empirical results show that our model outperforms the previous methods significantly on KB relation detection task and is expected to enable a KBQA system perform better than state-of-the-art KBQA systems.

\bibliography{reference}
\bibliographystyle{acl_natbib}

\end{document}